\documentclass{article}

\usepackage{microtype}
\usepackage{graphicx}
\usepackage{subcaption}
\usepackage{booktabs} 
\usepackage{multirow}
\usepackage{hyperref}

\usepackage[preprint]{icml2026}

\usepackage{ulem}
\usepackage{amsmath}
\usepackage{amssymb}
\usepackage{mathtools}
\usepackage{amsthm}
\usepackage{booktabs}
\usepackage[table]{xcolor}
\usepackage{amsmath}

\usepackage{subcaption}
\definecolor{myblue}{HTML}{71B7F7}
\definecolor{mypurple}{HTML}{C7AEF9}
\definecolor{myred}{HTML}{FF7C6E}

\usepackage{xcolor}

\usepackage[capitalize,noabbrev]{cleveref}

\theoremstyle{plain}

\theoremstyle{definition}

\theoremstyle{remark}

\newcommand{\swordsmanlogo}{%
  \raisebox{-0.3\height}{\includegraphics[height=3ex]{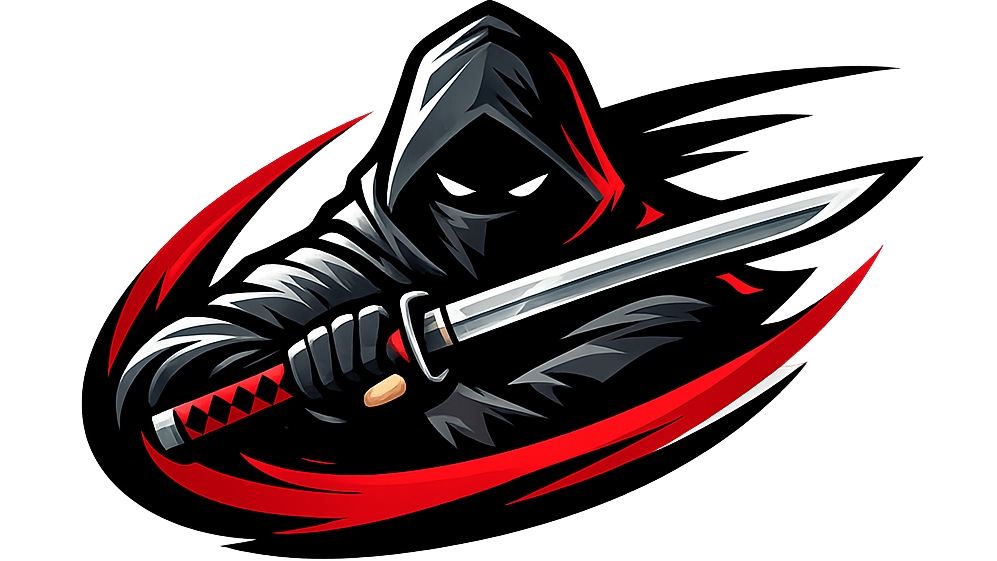}}%
  \hspace{0.35em}%
}

\usepackage[textsize=tiny]{todonotes}

\icmltitlerunning{Swordsman: Entropy-Driven Adaptive Block Partition for Efficient Diffusion Language Models}

\begin{document}

\twocolumn[
  \icmltitle{\texorpdfstring{\swordsmanlogo}{}Swordsman: Entropy-Driven Adaptive Block Partition\\for Efficient Diffusion Language Models}



  \icmlsetsymbol{equal}{*}
  \icmlsetsymbol{cor}{}

  \begin{icmlauthorlist}
    \icmlauthor{Yu Zhang}{equal,tj}
    \icmlauthor{Xinchen Li}{equal,tj}
    \icmlauthor{Jialei Zhou}{tj}
    \icmlauthor{Zhongwei Wan}{osu}
    \icmlauthor{Hongnan Ma}{uob}
    \icmlauthor{Yiwei Shi}{uob}
    \\
    \icmlauthor{Duoqian Miao}{tj,cor}
    \icmlauthor{Qi Zhang}{tj}
    \icmlauthor{Longbing Cao}{mqu}
  \end{icmlauthorlist}

  \icmlaffiliation{tj}{Tongji University}
  \icmlaffiliation{osu}{The Ohio State University}
  \icmlaffiliation{uob}{University of Bristol}
  \icmlaffiliation{mqu}{Macquarie University}

  \icmlcorrespondingauthor{Duoqian Miao}{dqmiao@tongji.edu.cn}

  \icmlkeywords{Machine Learning, ICML}

  \vskip 0.3in
]



\printAffiliationsAndNotice{}  

\begin{abstract}
Block-wise decoding effectively improves the inference speed and quality in diffusion language models (DLMs) by combining inter-block sequential denoising and intra-block parallel unmasking. However, existing block-wise decoding methods typically partition blocks in a rigid and fixed manner, which inevitably fragments complete semantic or syntactic constituents, leading to suboptimal performance. Inspired by the entropy reduction hypothesis (ERH), we recognize that constituent boundaries offer greater opportunities for uncertainty reduction, which motivates us to employ entropy analysis for identifying constituent boundaries. Therefore, we propose Swordsman, an entropy-driven adaptive block-wise decoding framework for DLMs. Swordsman adaptively partitions blocks by identifying entropy shifts between adjacent tokens to better align with semantic or syntactic constituent boundaries. In addition, Swordsman dynamically adjusts unmasking thresholds conditioned on the real-time unmasking status within a block, further improving both efficiency and stability. As a training-free framework, supported by KV Cache, Swordsman demonstrates state-of-the-art performance across extensive evaluations.
\end{abstract}

\section{Introduction}
In recent years, diffusion language models (DLMs)~\cite{austin2021structured, he2023diffusionbert, gong2022diffuseq} have rapidly emerged as a promising alternative to autoregressive language models by breaking the strictly sequential token decoding bottleneck through parallel decoding of multiple tokens at each denoising step. Large-scale DLMs such as LLaDA~\cite{nie2025large,zhu2025llada,zhu2025lladamoe}, Dream~\cite{ye2025dream}, Mercury~\cite{khanna2025mercury}, and Gemini Diffusion~\cite{deepmind2025geminidiffusion} have demonstrated the feasibility and scalability of the diffusion paradigm in large-scale language modeling. However, limited by diffusion's inherently numerous iterative denoising steps, DLMs currently face a severe challenge: while they possess a high theoretical efficiency ceiling, their actual inference speed and generation quality remain inferior to traditional autoregressive models in practice. This gap motivates the community actively explore better diffusion language decoding methods that balance both speed and quality ~\cite{li2025survey, zhong2026parallelism, hong2025wide}.

\begin{figure*}[t]
\centering
\includegraphics[width=\textwidth]{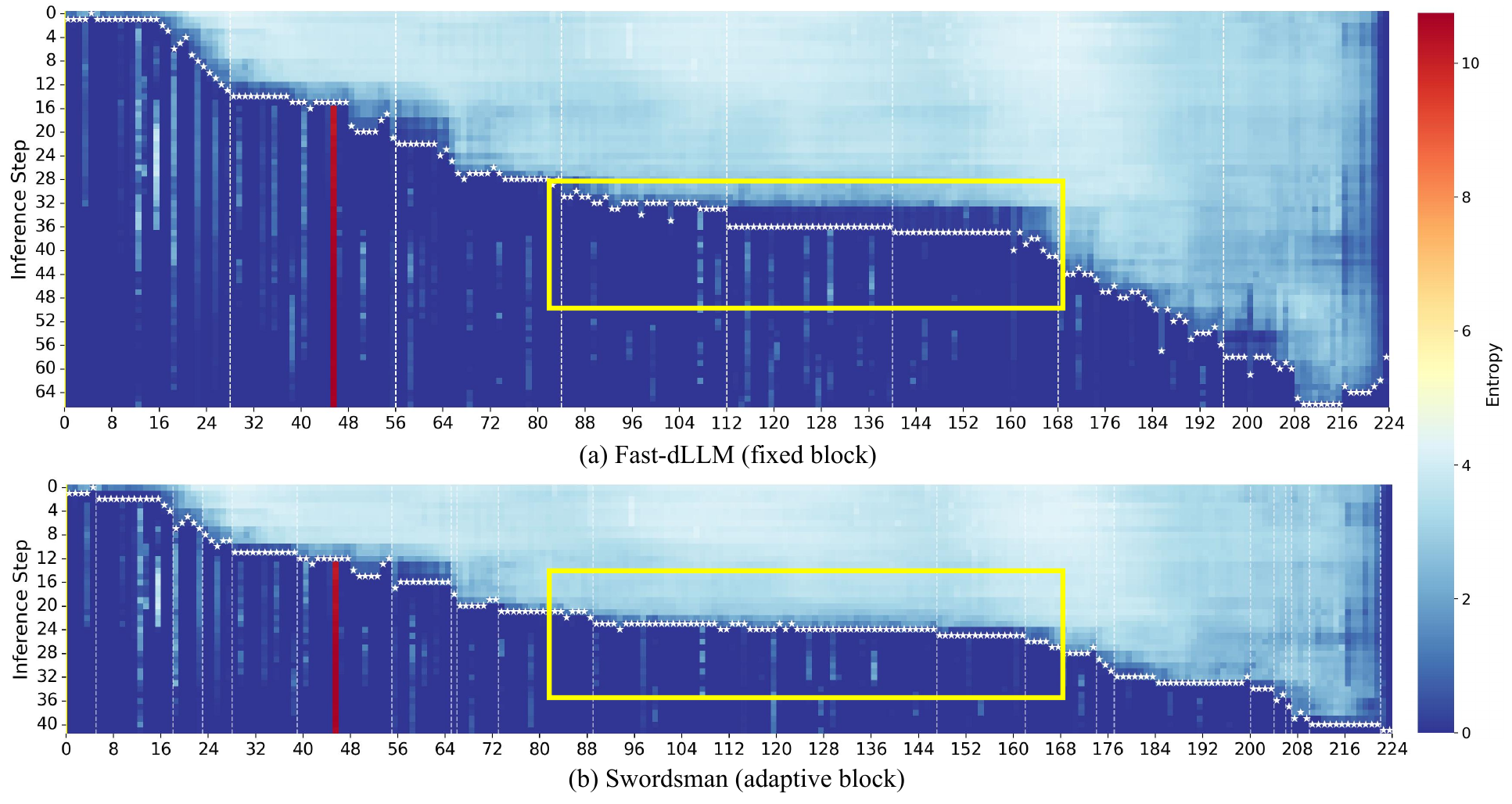}
\caption{Entropy evolution during decoding reveals semantic boundaries through shifts. Fast-dLLM (a) ignores entropy variance, applying predetermined fixed-length blocks that frequently fragment coherent constituents or merge semantically distinct ones, thereby degrading generation accuracy. Swordsman (b), however, leverages entropy shifts to adaptively align block boundaries with semantic constituents, partitioning at maximum shift points to achieve precise segmentation that yields better generations.}
\label{fig:introheatmap}
\end{figure*}

Block-wise decoding~\cite{wu2025fast,arriola2025block} has attracted widespread interest as a simple yet effective solution. It partitions the masked sequences into multiple decoding blocks to enable sequential inter-block denoising coupled with parallel intra-block decoding. Owing to their compatibility with the reuse of preceding blocks' KV cache~\cite{wu2025fastv2, wang2025diffusion, hu2025accelerating} and various scheduling strategies~\cite{huang2025pc,ben2025accelerated, hong2025wide}, block-wise decoding methods~\cite{wu2025fast,ma2025dinfer, chen2025dparallel} achieve an excellent trade-off between speed and quality under the same computational budget. Nevertheless, the introduction of blocks necessitates boundary partition, and the design of the boundary strategy directly impacts decoding performance. Existing boundary strategies typically adopt the fixed-length block partition, where the inherent rigidity and lack of adaptability often cause block boundaries to fail to align with semantic or syntactic constituents. This misalignment leads to tightly coupled tokens within the same constituent to be fragmented across different blocks, as illustrated in Figure~\ref{fig:introheatmap} (a), increasing token unmasking difficulty and ultimately degrading performance.

\begin{figure*}[h!]
\centering
\includegraphics[width=\textwidth]{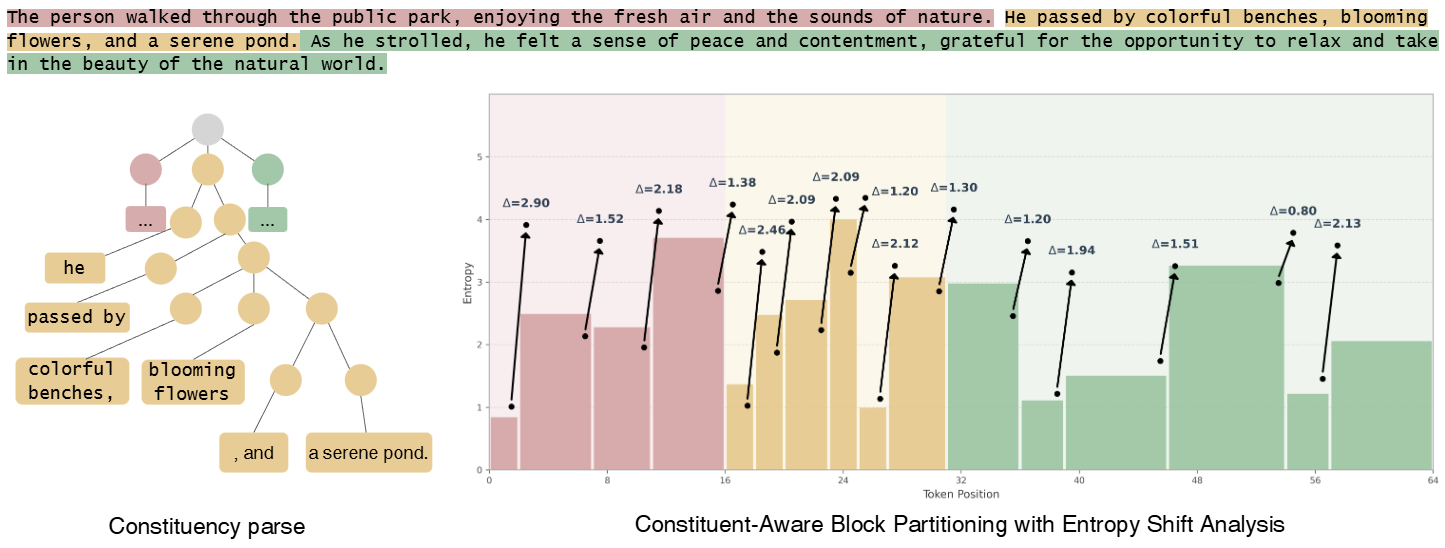}
\caption{Visualization of the alignment between entropy-driven block partitioning and the constituency-parse structure, showing that entropy-shift boundaries consistently match semantic constituents.}
\label{fig:entropy_shift}
\end{figure*}

This motivates a constituency-aware boundary strategy for block-wise decoding. Inspired by the entropy reduction hypothesis (ERH)~\cite{hale2006uncertainty}, we interpret the diffusion language generation process as progressively reducing uncertainty about subsequent semantic or syntactic continuations. The hypothesis posits that processing difficulty on a word is positively related to the entropy reduction it induces, i.e., uncertainty about the rest of the sentence after encountering that word. We extend this view to diffusion decoding: constituent boundaries correspond to shifts between syntactic states, and typically exhibit high uncertainty due to the vast freedom in perpetuating constituents. Take the sentence \texttt{I} \colorbox{myblue!50}{\texttt{DRINK}}  \colorbox{myred!50}{\texttt{ORANGE JUICE}} as an example. At the positions of constituents \colorbox{myblue!50}{\texttt{DRINK}} and \colorbox{myred!50}{\texttt{ORANGE JUICE}}, they admit a huge amount of alternatives for word choice, allowing for the replacement of with any verb or noun, respectively. Conversely, elements within a constituent are tightly coupled and represent continuous expansion of the same syntactic state, tending to exhibit smoother uncertainty evolution. Therefore, when processing words within constituents, the word alternative space (e.g., \colorbox{myred!50}{\texttt{()}} in \colorbox{myred!50}{\texttt{ORANGE()}}) is relatively narrow compared to boundary processing. Importantly, entropy serves as a direct measure of uncertainty. During diffusion decoding, the model produces token distributions at all positions, allowing us to directly calculate predictive entropy along the sequence. Meanwhile, as shown in Figure~\ref{fig:entropy_shift}, we conduct constituency parsing on selected sentences to obtain their constituents and analyze the entropy values at constituent boundaries. The analysis reveals the phenomenon wherein entropy variations generally exhibit large at constituent boundaries. Therefore, we advocate for adaptively partitioning constituent blocks based on the entropy shifts between adjacent tokens.

In light of the above discussion, we propose Swordsman, a training-free block-wise decoding framework for DLMs through entropy-driven adaptive block partition. Specifically, Swordsman calculates the predictive entropy of tokens during unmasking and then identifies constituent boundaries by detecting sharp entropy shifts between adjacent tokens to adaptively partition blocks. Furthermore, considering that adaptive block partition leads to variations in token unmasking scope and difficulty across blocks, Swordsman dynamically adjusts the unmasking thresholds based on the real-time status of parallel unmasking within the block to balance parallelism and reliability within each block.

Extensive experiments and analyses validate Swordsman as a simple yet effective block-wise decoding framework. Combined with KV Cache, Swordsman improves accuracy from 77.40\% to 81.50\% and achieves 8.79× faster inference on GSM8K compared to vanilla LLaDA. Compared with LLaDA-architecture Fast-dLLM, Swordsman achieves an accuracy improvement from 35.59\% to 43.90\% on Humaneval and delivers up to 8.31\% accuracy improvement while maintaining comparable inference speed. 

\section{Related Work}
\subsection{Diffusion Language Models}

Diffusion models~\cite{ho2020denoising}, as a generative paradigm based on iterative denoising, have been successfully applied in NLP as Diffusion Language Models (DLMs). Among DLMs, Masked Diffusion Models (MDMs)\cite{sahoo2024simple} emerge as the mainstream paradigm by surpassing alternatives\cite{austin2021structured, lou2310discrete} through simplified and efficient training objectives. Building upon MDMs, REMDM~\cite{wang2025remasking} enhances performance via a remasking sampler for iterative inference refinement, while A-CFG~\cite{li2025adaptive} balances generation diversity and prompt fidelity through dynamic low-confidence masking. However, these methods fail to address critical bottlenecks in inference latency and generation quality. Inspired by scaling laws, frameworks like LLaDA have further unlocked DLM potential through architectural adjustments and parameter scaling, becoming leading foundation models.

\subsection{Efficient inference for DLMs}
Recently, researchers have improved LLaDA from various perspectives. FlexMDM~\cite{kim2025any} addresses variable-length generation through flexible-order masked diffusion. DCoLT~\cite{huang2025reinforcing} enhances reasoning capabilities via chain-of-thought methods. LLaDA 1.5~\cite{zhu2025llada} incorporates VRPO-based reinforcement learning for better human preference alignment. Despite progress in specific areas, these methods fail to address LLaDA's core challenges: high inference latency and suboptimal generation quality, limiting its broader applications.

Several methods have been proposed to address these challenges. PC-Sampler~\cite{huang2025pc} improves quality by adjusting token decoding order. APD~\cite{israel2025accelerating} accelerates inference via a small autoregressive model. However, these methods rely on bidirectional attention, imposing substantial computational overhead that limits practical inference latency even with parallelization.

Thus, block-wise decoding~\cite{wu2025fast,wu2025fastv2,arriola2025block} has emerged to address these efficiency concerns by organizing text into blocks. Decoded blocks can provide cached information for subsequent blocks, thereby accelerating inference~\cite{kim2025cdlm, wang2025diffusion}. Meanwhile, block-by-block decoding prevents premature generation of unnatural content (i.e., [EOF])\cite{zhu2025llada}. With advantages in both latency and quality, block-wise methods have become the mainstream solution. However, existing methods typically use fixed block sizes~\cite{hong2025wide, agrawal2025spiffy}, ignoring semantic constituents within sentences.  This causes abrupt constituent boundaries and internal confusion within blocks. Current preliminary explorations attempt to partition blocks using punctuation marks\cite{lu2025adablock}, but such coarse-grained segmentation offers limited efficiency improvement. To address this, we develop a training-free fine-grained partitioning framework that organizes blocks by semantic constituents, maintaining low inference latency while improving generation quality.

\section{Methodology}
\subsection{Preliminary}
\textbf{Decoding Process of DLMs.} Diffusion Language Models (DLMs) generate text through a masked diffusion paradigm~\cite{nie2025large}. Starting from a fully masked sequence $\mathbf{x}^{(T)}$ where all tokens are [MASK], the model progressively unmasks tokens over $T$ steps. At each step $t \in \{T, T-1, \dots, 1\}$ with the updated sequence $\mathbf{x}^{(t)}$, the decoding process is modeled as:
\begin{equation} 
\vspace{-5pt}
p_\theta(\mathbf{x}^{(t-1)} | \mathbf{x}^{(t)}) = \prod_{i \in \mathcal{M}(t)} p_\theta(x_i | \mathbf{x}^{(t)}), 
\end{equation}
where $\mathcal{M}(t)$ contains masked positions at step $t$ and $p_\theta$ denotes the learned decoding transition. Common confidence-based samplers unmask tokens using threshold $\tau$:
\begin{equation}
\begin{aligned}
    &c_i^{(t)} \triangleq \max_{v \in \mathcal{V}} p_\theta(v \mid \mathbf{x}^{(t)}, i), \\
    &\mathcal{U}^{(t)} = \left\{ i \in \mathcal{M}^{(t)} \mid c_i^{(t)} \geq \tau \right\}, \\
\end{aligned}
\end{equation}
where $\mathcal{V}$ denotes the vocabulary, $c_i^{(t)}$ is the confidence score of token $i$, and $\mathcal{U}^{(t)}$ represents the set of tokens selected for unmasking at step $t$.
The sequence $\mathbf{x}^{(t)}$ is then updated as:
\begin{equation}
    x_i^{(t-1)} = 
    \begin{cases} 
        \operatorname*{argmax}\limits_{v \in \mathcal{V}} p_\theta(v \mid \mathbf{x}^{(t)}, i), & \text{if } i \in \mathcal{U}^{(t)}, \\
        x_i^{(t)}, & \text{otherwise},
    \end{cases}
\end{equation}
with the masked set updated as $\mathcal{M}^{(t-1)} = \mathcal{M}^{(t)} \setminus \mathcal{U}^{(t)}$. This iterative process continues until the masked set is empty, i.e., $\mathcal{M}^{(t)} = \emptyset$, resulting in the final generated sequence $\mathbf{x}^{(0)}$.

\textbf{Block-wise DLMs.} Block-wise decoding~\cite{wu2025fast, arriola2025block} partitions the sequence into $K$ disjoint blocks, denoted as $\mathcal{B} = \{B_1, B_2, \dots, B_K\}$ and decodes them sequentially. At each step $t$, only tokens within the current block $B^{(t)}$ are considered for unmasking: 
\begin{equation}
S^{(t)}= \{i \mid i \in B^{(t)}\},
\end{equation}
where $S^{(t)}$ constricts decoding process to the right positions, preventing premature unnatural content generation for better generation quality while enabling two key mechanisms: KV cache across sequential blocks and parallel decoding within each block, both accelerating inference.

\subsection{Entropy Analysis}
Entropy measures the uncertainty of a probability distribution in information theory~\cite{mackay2003information}. For DLMs, given context $\mathcal{C}_i$, the entropy of the $i^{th}$  token $t_i$ is defined as the Shannon entropy of the predictive distribution:
\begin{equation}
\label{eq:entropy}
H_i = -\sum_{v \in \mathcal{V}} p_\theta(t_i = v \mid \mathcal{C}_i) \log p_\theta(t_i = v \mid \mathcal{C}_i),
\end{equation}
where $v \in \mathcal{V}$ denotes a candidate token. We utilize the entropy shift $\Delta H_i = H_{i+1} - H_i$ to capture the dynamic changes of uncertainty. Formally, a text sequence $\mathbf{x}$ is composed of multiple independent semantic constituents $\{\mathcal{G}_1, \mathcal{G}_2, \dots, \mathcal{G}_K\}$. Tokens within constituents are tightly coupled syntactically and semantically, forming strong correlations, while different constituents exhibit weak dependencies due to their distinct semantic expressions.

In DLMs, entropy quantifies the scale of the effective token search space. Under a uniform distribution assumption over the candidate vocabulary $V_{\text{eff}}(i)$ of $t_i$, with the size denoted as $N_i = |V_{\text{eff}}(i)|$, the entropy of $t_i$ can be modeled as:
\begin{equation}
\label{eq:vocabulary}
H_i \approx \log N_i + \epsilon_{\text{flat}},
\end{equation}
where $\epsilon_{\text{flat}}$ is a shape correction coefficient. According to Equation~\ref{eq:entropy}, the entropy shift $\Delta H_i$ is determined by the ratio of candidate vocabulary sizes of adjacent tokens as:
\begin{equation}
\label{eq:spike}
\Delta H_i = H_{i+1} - H_i \approx \log \frac{N_{i+1}}{N_i},
\end{equation}
establishing vocabulary size differences as the primary determinant of entropy shifts.

\textbf{Intra-constituent Smoothness.} Within a semantic constituent, adjacent positions share consistent syntactic and semantic constraints from the decoded prefix. Since they belong to the same incomplete semantic constituent, the candidate vocabulary size exhibits smooth variation. We formalize this as the \textbf{Intra-Constituent Smoothness Assumption}: for adjacent token $t_i, t_{i+1} \in \mathcal{G_A}$, the relative change in candidate vocabulary size satisfies:
\begin{equation}
    \left| \frac{N_{i+1} - N_i}{N_i} \right| = |\xi| \leq \delta, ~~\delta \in [0, 1),
\end{equation}
where $\delta$ is a constant representing the local prediction uncertainty. When $\delta$ is small, applying the Taylor expansion for $\log(1+x) \approx x$ yields the upper bound of entropy shift within the same constituent as:
\begin{equation}
|\Delta H_i^{\text{intra}}| \approx |\log(1 + \xi)| \approx |\xi| \le \delta.
\end{equation}
This proves that within semantic constituents, entropy shifts are constrained to a small neighborhood.

\textbf{Constituent Boundary Abruptness.} When decoding transitions from position  $i \in \mathcal{G}_A$ to $i+1 \in \mathcal{G}_B$, the local strong constraints imposed by $\mathcal{G}_A$ terminate, while those of $\mathcal{G}_B$ remains unestablished. The prediction space degrads from specific constituent expression $N_{\text{local}}$ to general semantic text $N_{\text{global}}$, where the ratio $\rho$ meets:

\begin{equation}
\rho = \frac{N_{\text{global}}}{N_{\text{local}}} \gg 1.
\end{equation}
With Equation~\ref{eq:vocabulary}, the boundary entropy shift can be put as:
\begin{equation}
\Delta H_i^{\text{boundary}} = H_{i+1}-H_i\approx \log \frac{N_{\text{global}}}{N_{\text{local}}}=\log \rho.
\end{equation}
Since $N_{\text{global}}$ spans a wide range of syntactic and semantic contexts, it's clear that $\rho$ is substantially large.

Combining the above analysis, the feasibility of semantic boundary detection depends on the following ratio:
\begin{equation}
\frac{\Delta H_i^{\text{boundary}}}{|\Delta H_i^{\text{intra}}|} \approx \frac{\log \rho}{\delta},
\end{equation}
when semantic constituents maintain internal coherence ($\delta$ small) and exhibit distinct semantic domains($\rho$ large), we have $\Delta H_i^{\text{boundary}} \gg |\Delta H_i^{\text{intra}}|$. Therefore, detecting local maxima of entropy shifts enables accurate identification of semantic constituent boundaries.

\subsection{Swordsman}
\begin{figure}[htbp]
\centering
\includegraphics[width=\linewidth]{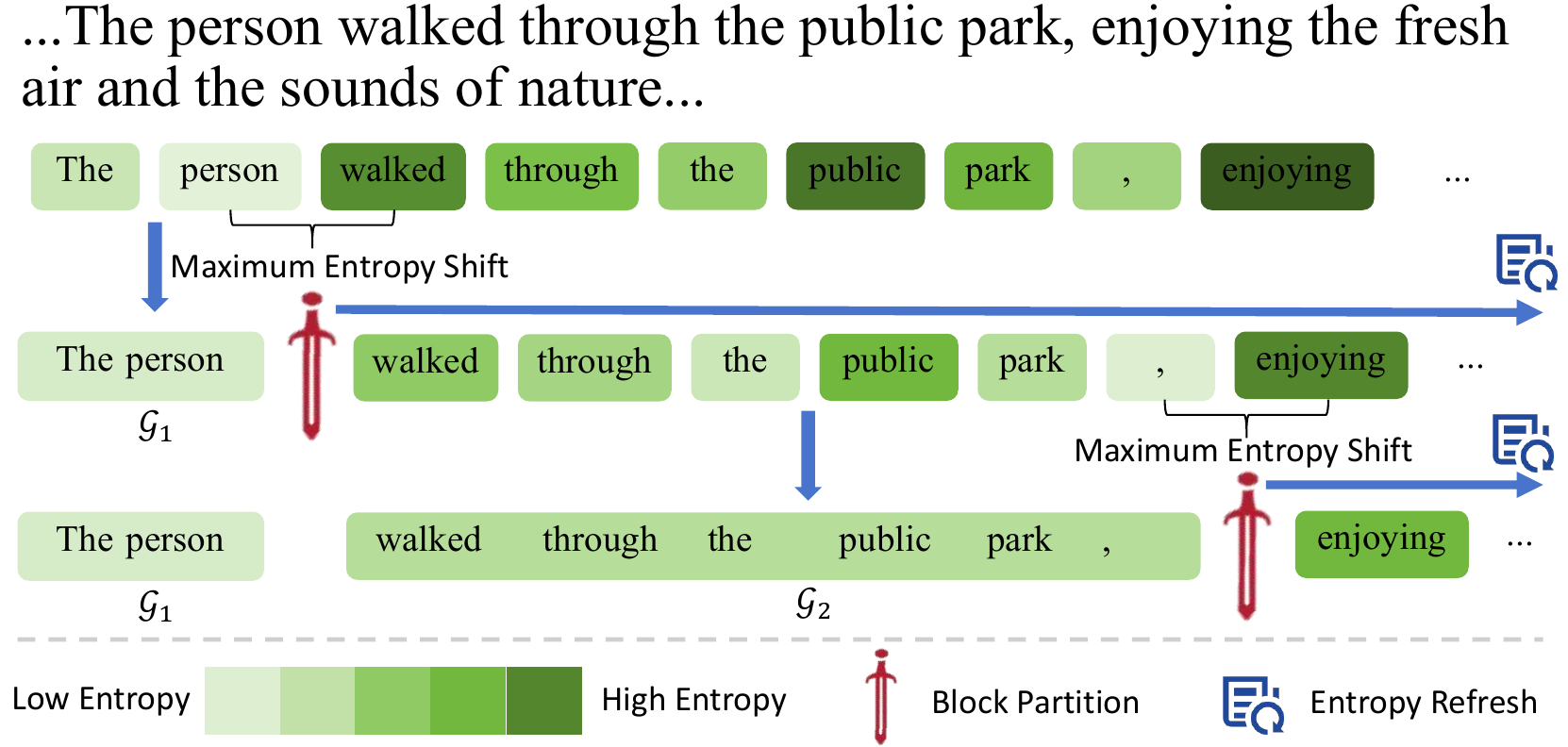}
\caption{Entropy-driven adaptive partitioning for semantic constituents: each step splits at the maximum entropy shift, decodes the block with a dynamic threshold, refreshes entropy, and repeats.} 
\label{framework}
\end{figure}

Building upon the entropy analysis in Section 3.2, we present Swordsman, a training-free block-wise decoding framework that leverages entropy shifts to achieve constituent-aware generation in DLMs. As illustrated in Figure~\ref{framework}, Swordsman employs entropy-driven adaptive partitioning to align block boundaries with semantic constituents during generation. Subsequently, each partitioned block is decoded via difficulty-aware parallel unmasking with dynamic thresholds tailored to block-specific uncertainty.

\paragraph{Adaptive block partition}

In the text sequence generation task of DLMs, block-wise decoding models inherently require partitioning the sequence into blocks to enable progressive, block-by-block generation. This partitioning is not merely a technical detail but fundamentally determines the quality of decoding: improper boundaries split semantic constituents, significantly degrading generation performance. In contrast to traditional methods that employ rigid partitioning with predetermined fixed-length blocks before inference, we propose adaptive partitioning that divides block boundaries based on the local maximum entropy shifts $\Delta H_i$ between adjacent tokens during decoding, which reveal the natural transitions between semantic constituents. 

Specifically, we employ an iteratively partitioning strategy that progressively determines block boundaries once prefix blocks are decoded. At each iteration after the  $k^{th}$ block $B_k$ is decoded, the previously decoded content $\{B_1,…,B_k\}$ provides essential contextual information that refines the model's uncertainty estimates for the remaining masked positions. We re-perform forward propagation on the residual sequence to compute the updated predictive entropy $H_i$ at each remaining position, calculate the entropy shifts $\Delta H_i$ based on these updated values, and identify the next maximum shift point as the right boundary for current block: 
\begin{equation}
    position_r = \begin{cases} 
        \operatorname*{argmax}\limits_{t < i \le L} \Delta H_i & \text{if } \max\limits_{t < i \le L} \Delta H_i \ge \tau_{\min}, \\
        L & \text{otherwise},
    \end{cases}
\end{equation}
where $L$ denotes the predefined generation length. As decoding progresses and contextual information accumulates, the uncertainty of remaining tokens diminishes, reflected in smaller entropy shifts. To prevent inefficient over-segmentation in the tail region where uncertainty has substantially converged, we introduce an adaptive termination mechanism with a minimum shift threshold $\tau_{min}=0.1$, below the entropy shift at semantic constituent boundaries. When the maximum entropy shift in the remaining sequence falls below this threshold, we terminate further partitioning and merge all left tokens into the current block, thereby avoiding unnecessary computational overhead from excessive partitioning while maintaining decoding efficiency.

This entropy-driven partitioning yields two key advantages: intra-block semantic consistency reduces prediction uncertainty within constituents, while inter-block semantic independence allows efficient KV cache reuse without costly out-of-block re-computations. These properties jointly contribute to improved generation quality and reduced inference latency relative to fixed partitioning strategies.

\vspace{-5pt}
\paragraph{Dynamic parallel unmask} 
Due to the precise semantic-based partitioning, tokens within each block are tightly coupled in syntax and semantics, while blocks exhibit significant variations in both scope and difficulty. 
However, existing static decoding schemes apply a constant confidence threshold across all blocks. This causes severe mismatches: strict thresholds force unnecessary serial decoding in confident blocks, while lenient thresholds enable risky parallelization in uncertain blocks. Such difficulty-agnostic threshold settings ignore the collaboration of tokens within semantic constituents and fail to flexibly adapt decoding strategies to varying block difficulty.

To further leverage the acceleration benefits brought by semantic priors, we propose a dynamic threshold unmask mechanism based on block average entropy as put:
\begin{equation}
    \bar{H}_k = \frac{1}{|B_k|} \sum_{t_i \in B_k} H_i.
\end{equation}
Before decoding each block, we obtain the overall uncertainty of the $k^{th}$ block $B_k$ by calculating its initial average entropy $\bar{H}_k$. High entropy indicates semantic complexity requiring cautious decoding, while low entropy enables aggressive parallelization in the block decoding. This motivates our dynamic threshold unmask mechanism: adapting the confidence threshold to block-specific difficulty enables both safe decoding for uncertain blocks and aggressive parallelization for confident blocks.

Furthermore, we calibrate the $k^{th}$ block's difficulty using historical entropy from preceding blocks as: 
\begin{equation}
    \lambda_k = 1 - \frac{\bar{H}_k}{\bar{\mathcal{H}}_{max}^{(k)}}, ~~\bar{\mathcal{H}}_{max}^{(k)} = \max_{i=1}^k \bar{H}_i~,
\end{equation}
Based on the calibrated difficulty metric $\lambda_k$, we compute the dynamic confidence threshold for the block $B_k$ as:
\begin{equation}
    \tau_t = \tau_{init} \cdot \left[ (1 - \lambda_k) + \lambda_k \cdot \sqrt{\frac{\bar{H}_t^{(k)}}{\bar{H}_{start}^{(k)}}} \right],
\end{equation}
where $\tau_{init}$ denotes the global base threshold, while $\bar{H}_t^{(k)}$ and $\bar{H}_{start}^{(k)}$ represents the average entropy of block $B_k$ at the current decoding step $t$ and the initial step respectively. 

This dynamic mechanism enables two-level adaptive thresholding. At the inter-block level, the coefficient $\lambda_k$ modulates the threshold based on block-specific difficulty. High-entropy blocks use stricter thresholds to ensure accuracy, while low-entropy blocks use relaxed ones to maximize parallelization. At the intra-block level, the ratio $\sqrt{\frac{\bar{H}_t^{(k)}}{\bar{H}_{start}^{(k)}}}$ captures the entropy decay as decoding progresses within the block. As more tokens are decoded, $\bar{H}_t^{(k)}$ decreases, causing $\tau_t$ to gradually decline. This progressive relaxation allows the model to perform strict selection initially, but gradually ease constraints for better parallelization.

These two levels of threshold adaptation work complementarily: inter-block thresholding matches decoding strategies to semantic constituents' difficulties, while intra-block adjustment dynamically exploits decreasing uncertainty within each constituent.
This maximizes inference acceleration while ensuring generation quality, achieving a dynamic balance between speed and quality.

\section{Experiments}

\subsection{Experimental Setup}
\textbf{Implementation Details} Our method is evaluated on three pretrained DLMs: LLaDA-8B-Instruct \cite{nie2025large}, LLaDA-1.5 \cite{zhu2025llada}, and Dream-v0-Base-7B \cite{ye2025dream}. All experiments are conducted on an NVIDIA H200 GPU. We compare our proposed method against three baselines, including two fixed block methods: \textbf{Fast-dLLM} and \textbf{D2F}~\cite{wang2025diffusion}; and one adaptive block partitioning strategy \textbf{AdaBlock}~\cite{lu2025adablock}. Since AdaBlock is not publicly released, we compare against the results reported in its paper, whereas the other open-sourced baselines are reproduced locally under the same experimental setting. For all tasks, we set the generation budget length to $L = 512$, and the static confidence threshold $\tau_{fixed} = 0.9$. Swordsman introduces two hyperparameters: the minimum entropy shift $\tau_{min}=0.1$ of the adaptive block partition strategy and the base threshold $\tau_{init}=0.9$ of dynamic parallel unmask mechanism. For fixed block methods, the block size is set to 32 (the optimal set for speed and accuracy identified by Fast-dLLM). Furthermore, we evaluate the methods across three KV cache configurations: no cache, prefix cache, and dual caches.

\textbf{Benchmarks and Metrics.} We evaluate on standard LLM benchmarks covering two categories: 
(1) code generation, including HumanEval (0-shot) and MBPP (3-shot); 
(2) mathematical reasoning, including GSM8K (5-shot) and MATH (4-shot). 
For evaluation metrics, we report \textbf{Accuracy} (\%, $\Uparrow$) 
for generation quality, \textbf{Throughput} (TPS, $\Uparrow$) for 
decoding speed, and \textbf{Latency} (s/sample, $\Downarrow$) for end-to-end generation time. Throughout the tables, \textbf{bold} and 
\underline{underline} denote the best and second-best results, respectively. 

\subsection{Main Results}

\paragraph{Overall Performance}
Table~\ref{tab:results} presents the comprehensive comparison across three DLMs and four benchmarks. Swordsman achieves \textbf{state-of-the-art} performance across multiple models and benchmarks, validating the efficiency of our entropy-driven partitioning approach. Specifically, on LLaDA-8B-Instruct, Swordsman achieves 81.50\% on GSM8K, surpassing Fast-dLLM by +6.29\% and AdaBlock by +3.00\%. On LLaDA-1.5, Swordsman attains the best performance across all metrics, notably reaching 84.00\% on GSM8K (+1.44\%) and 43.90\% on HumanEval (+8.31\% over Fast-dLLM). For Dream-v0-base-7B, Swordsman yields the highest HumanEval accuracy of 57.93\%, outperforming all baselines. These improvements validate our core hypothesis: entropy-driven adaptive partitioning better captures the natural structure of semantic constituents, enabling more efficient parallel decoding. We observe that AdaBlock reports unusually high MBPP scores on LLaDA-8B-Instruct with low performance on Dream-v0-Base-7B, likely due to evaluation protocol and post-processing differences.

\begin{table}[t]
\centering
\caption{Generation quality comparison of different models with various cache configurations on four benchmarks.}
\label{tab:results}
\setlength{\tabcolsep}{4pt}
\renewcommand{\arraystretch}{1}
\resizebox{\linewidth}{!}{
\begin{tabular}{lllcccc}
\toprule
\multicolumn{3}{c}{\textbf{Experiment Setting}} & \multicolumn{4}{c}{\textbf{Accuracy} $\Uparrow$} \\
\cmidrule(lr){1-3}\cmidrule(lr){4-7}
\textbf{Partition} & \textbf{Method} & \textbf{Cache} & \textbf{GSM8K} & \textbf{MATH} & \textbf{Humaneval} & \textbf{MBPP} \\
\midrule

\rowcolor{gray!15}
\multicolumn{7}{c}{\textbf{LLaDA-8B-Instruct}} \\
\multirow{4}{*}{Fixed}
  & \multirow{3}{*}{Fast-dLLM} & None   & 77.56 & 36.52 & 43.90 & 14.20 \\
  &                           & Prefix & 77.10 & 36.22 & 41.46 & 13.20 \\
  &                           & Dual   & 75.21 & 35.46 & \underline{44.51} & 13.60 \\
\cmidrule(lr){2-7}
  & D2F                       & Prefix & 74.98 & 28.76 & 36.59 & \underline{38.00} \\
\cmidrule(lr){1-7}
\multirow{5}{*}{Adaptive}
  & \multirow{2}{*}{AdaBlock} & None   & 80.60 & \textbf{37.30} & 43.30 & \textbf{39.80} \\
  &                           & Dual   & 78.50 & 35.30 & \textbf{46.30} & \underline{38.00} \\
\cmidrule(lr){2-7}
  & \multirow{3}{*}{Swordsman}& None   & \underline{81.43} & \underline{36.82} & 42.68 & 13.00 \\
  &                           & Prefix & 80.67 & 36.68 & 43.90 & 13.60 \\
  &                           & Dual   & \textbf{81.50} & 35.76 & \underline{44.51} & 13.60 \\

\midrule
\rowcolor{gray!15}
\multicolumn{7}{c}{\textbf{Dream-v0-base-7B}} \\
\multirow{4}{*}{Fixed}
  & \multirow{3}{*}{Fast-dLLM} & None   & 75.97 & \textbf{40.08} & 52.44 & 55.40 \\
  &                           & Prefix & 74.00 & 38.96 & \underline{56.70} & \textbf{55.80} \\
  &                           & Dual   & 74.75 & 38.34 & 53.05 & 54.40 \\
\cmidrule(lr){2-7}
  & D2F                       & Prefix & 76.12 & 38.62 & 52.55 & 53.48 \\
\cmidrule(lr){1-7}
\multirow{5}{*}{Adaptive}
  & \multirow{2}{*}{AdaBlock}  & None   & 75.70 & 39.90 & 51.20 & 14.20 \\
  &                            & Dual   & 75.10 & 38.40 & 53.00 & 11.60 \\
\cmidrule(lr){2-7}
  & \multirow{3}{*}{Swordsman} & None   & 75.82 & \underline{40.00} & 54.27 & \underline{55.60} \\
  &                            & Prefix & \textbf{76.88} & 39.42 & \textbf{57.93} & \textbf{55.80} \\
  &                            & Dual   & \underline{76.50} & 38.58 & 55.49 & 54.80 \\

\midrule
\rowcolor{gray!15}
\multicolumn{7}{c}{\textbf{LLaDA-1.5}} \\
\multirow{3}{*}{Fixed}
  & \multirow{3}{*}{Fast-dLLM} & None   & 82.56 & \textbf{37.18} & 39.02 & 39.60 \\
  &                           & Prefix & 81.80 & 34.60 & 39.02 & \underline{40.00} \\
  &                           & Dual   & 80.52 & 33.26 & 35.59 & 36.20 \\
\cmidrule(lr){1-7}
\multirow{5}{*}{Adaptive}
  & \multirow{2}{*}{AdaBlock}  & None   & 82.40 & 36.70 & 38.40 & 37.60 \\
  &                            & Dual   & 81.70 & 33.90 & 39.00 & 36.40 \\
\cmidrule(lr){2-7}
  & \multirow{3}{*}{Swordsman} & None   & \textbf{84.00} & 36.58 & \underline{42.68} & \textbf{41.00} \\
  &                            & Prefix & 82.56 & \underline{36.94} & 42.02 & 39.80 \\
  &                            & Dual   & \underline{82.87} & 35.30 & \textbf{43.90} & 39.40 \\

\bottomrule

\end{tabular}
}
\end{table}

\vspace{-8pt}
\paragraph{Analysis of Inference Speed Performance}

We further analyze the inference speed performance on GSM8K across DLMs, as shown in Table~\ref{tab:efficiency}. Under the same caching configuration, Swordsman consistently matches or exceeds Fast-dLLM throughput while maintaining the best accuracy. Across all three backbones, Swordsman consistently delivers top-tier inference speed, achieving the best or near-best performance in throughput and latency metrics. Under the strongest cache configuration (Dual), Swordsman increases TPS over Fast-dLLM by +1.62 (LLaDA-8B-Instruct), +1.48 (Dream-v0-base-7B), and +3.67 (LLaDA-1.5), averaging +2.26 TPS, while simultaneously reducing latency by -0.31s. The most pronounced improvement occurs on LLaDA-1.5, where Swordsman achieves both the highest TPS (64.97) and the lowest latency (3.03s), demonstrating a clear speed advantage. While D2F achieves higher TPS on Dream-7B (89.82) by distilling multi-step reasoning into single-step inference during training, it suffers significant accuracy degradation as shown in Table~\ref{tab:results}. In contrast, Swordsman operates as a training-free framework that achieves substantial speedup purely through adaptive block partitioning, dynamic parallel unmasking, and caching strategies. These results demonstrate that Swordsman successfully balances speed and quality without requiring model retraining, offering a more practical approach that maintains both competitive inference speed and superior accuracy.

\begin{table}[t]
\centering
\caption{Inference speed comparison of different models with various cache configurations on four benchmarks.}
\label{tab:efficiency}
\setlength{\tabcolsep}{3.0pt}
\renewcommand{\arraystretch}{0.92}
\resizebox{\linewidth}{!}{
\begin{tabular}{llcccccc}
\toprule
& & \multicolumn{2}{c}{\textbf{LLaDA-8B-Instruct}} & \multicolumn{2}{c}{\textbf{Dream-v0-base-7B}} & \multicolumn{2}{c}{\textbf{LLaDA-1.5}} \\
\cmidrule(lr){3-4}\cmidrule(lr){5-6}\cmidrule(lr){7-8}
\textbf{Method} & \textbf{Cache} & TPS $\Uparrow$ & Latency $\Downarrow$ & TPS $\Uparrow$ & Latency $\Downarrow$ & TPS $\Uparrow$ & Latency $\Downarrow$ \\
\midrule
\multirow{3}{*}{Fast-dLLM} 
  & None   & 42.36 & 6.61  & 39.98 & 12.54 & 40.43 & 5.23 \\
  & Prefix & 58.42 & 4.59  & 62.75 & 8.21  & 50.16 & 3.90 \\
  & Dual   & \underline{72.12} & 3.72  & 74.37 & 7.11  & \underline{61.30} & 3.41 \\
\midrule
D2F & Prefix & 64.49 & \textbf{2.24} & \textbf{89.82} & \textbf{4.32} & -- & -- \\
\midrule
\multirow{2}{*}{AdaBlock} 
  & None   & 45.02 & 5.98  & 43.12 & 11.64 & 41.06 & 4.92 \\
  & Dual   & 58.21 & 4.73  & 63.73 & 8.14  & 54.69 & \underline{3.16} \\
\midrule
\rowcolor{yellow!12}
\cellcolor{yellow!12} & None   & 44.89 & 6.00 & 42.34 & 12.07 & 41.14 & 4.89 \\
\rowcolor{yellow!12}
\cellcolor{yellow!12}\multirow{-2}{*}{\textbf{Swordsman}} & Prefix & 58.68 & 4.54 & 63.88 & 7.99  & 52.55 & 3.69 \\
\rowcolor{yellow!12}
\cellcolor{yellow!12} & Dual   & \textbf{73.74} & \underline{3.66} & \underline{75.85} & \underline{6.74}  & \textbf{64.97} & \textbf{3.03} \\
\bottomrule
\end{tabular}
}
\end{table}

\paragraph{Robustness across Cache Configurations} Despite the fact that caching strategies introduce computational approximations that can degrade accuracy, as evidenced by Fast-dLLM's performance drop from 77.56\% (None) to 75.21\% (Dual) on GSM8K for LLaDA-8B-Instruct, Swordsman maintains consistently superior performance across all cache configurations, delivering stable gains over Fast-dLLM. Aggregating results across the four benchmarks, Swordsman surpasses Fast-dLLM under every cache setting with an average improvement of about +1.50\% and the advantage becomes more pronounced when stronger caching is used, specifically on HumanEval under Dual caching for LLaDA-1.5, where Swordsman achieves 43.90\% but 35.59\% for Fast-dLLM, an absolute gain of +8.31\%, highlighting Swordsman’s markedly higher robustness to cache-induced approximation in code-generation settings. Across the entire datasets, Swordsman consistently achieves substantial speedup from the caching strategy according to Table~\ref{tab:efficiency} compared to the non-caching performance.

\paragraph{Analysis of Efficiency}

\begin{figure}[htbp] 
    \centering 
    \includegraphics[width=\linewidth]{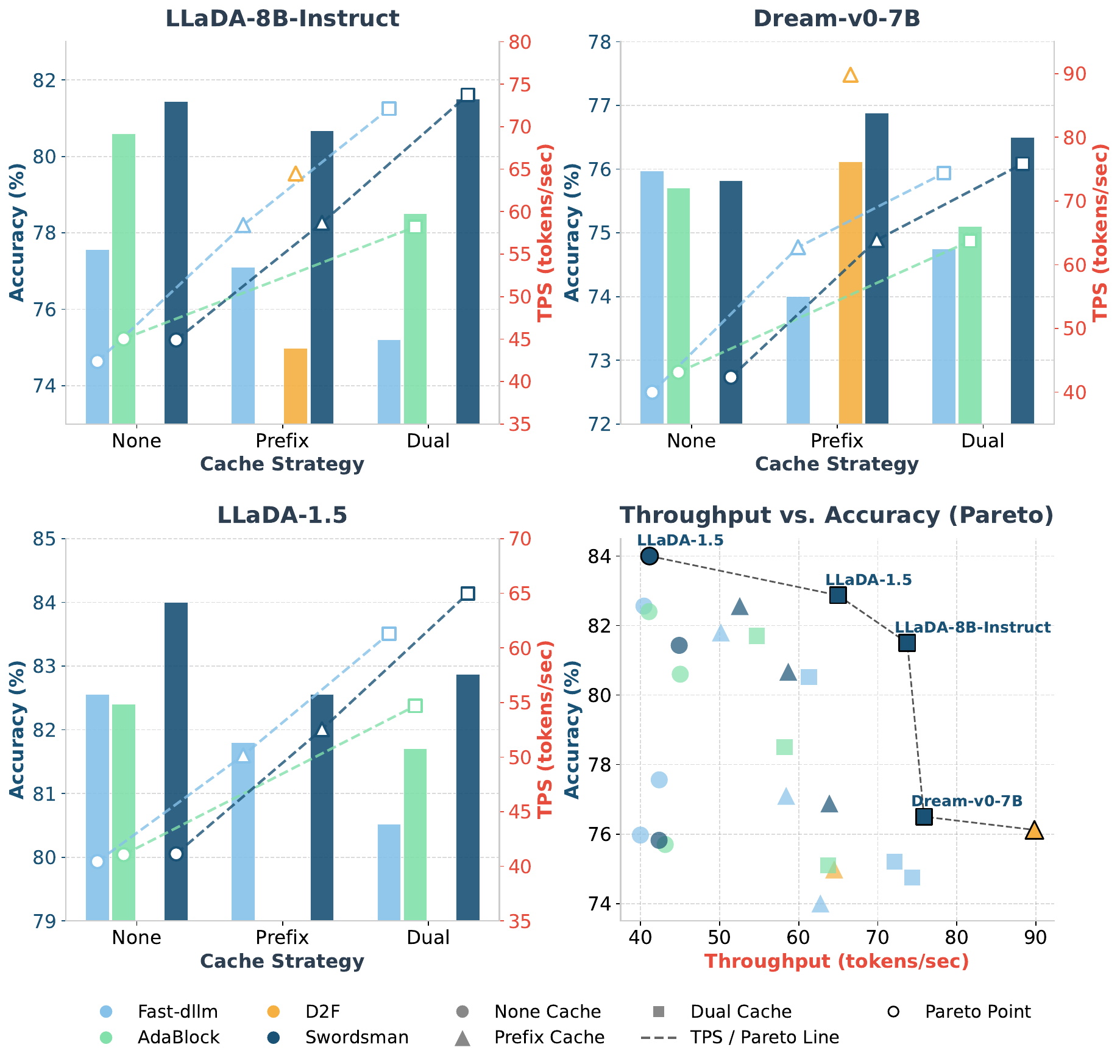} 
    \caption{Trade-off between throughput and accuracy} 
    \label{throughput_accuracy}
\end{figure}

Figure \ref{throughput_accuracy} compares efficiency across cache configurations and backbones. Swordsman consistently delivers the best or near-best accuracy across all settings, with stable gains over Fast-dLLM. On GSM8K, it reaches 81.50\% on LLaDA-8B-Instruct under Dual cache (vs. 75.21\% for Fast-dLLM) and attains the overall peak of 84.00\% on LLaDA-1.5 with None cache, while maintaining 82.87\% under Dual. Crucially, these accuracy gains do not sacrifice speed. Averaged over all comparable settings on GSM8K, Swordsman yields +1.79 higher throughput and -0.3s lower latency than Fast-dLLM, with a {+2.53\%} accuracy improvement. Under Dual caching specifically, it achieves a larger gain of {+2.20 TPS, -0.27s latency, and +3.46\% accuracy} with Fast-dLLM.
This joint superiority is clearly reflected in the Pareto analysis: four Swordsman configurations lie on the frontier, each achieving superior speed-accuracy combinations. While D2F also reaches the frontier through training-based distillation, it occupies the extremely high-speed, low-accuracy region. The strongest demonstration of Swordsman's superiority is on LLaDA-1.5 with Dual cache, simultaneously achieving high throughput (64.97 TPS), the lowest latency (3.03s), and strong accuracy (82.87\%), demonstrating that entropy-driven block partitioning and dynamic threshold adjustment for parallel unmasking enable both high inference speed and superior generation quality without model retraining.

\paragraph{Analysis of Decoding Strategies} Table~\ref{decodingablation} presents a decoing strategy analysis on LLaDA-8B-Instruct with GSM8K. Starting from full-diffusion LLaDA (29.04\%, 3.32 TPS), introducing block-wise decoding dramatically improves accuracy to 77.40\% (+48.36\%) with 2.5$\times$ speedup (8.39 TPS) by enforcing left-to-right generation order. Adding parallel decoding by setting a confidence threshold for token selection yields a further 5$\times$ speedup (8.39$\rightarrow$42.36 TPS) with negligible accuracy change. Finally, our adaptive partitioning boosts accuracy to 81.43\% (+3.87\%) by aligning block boundaries with semantic constituents via entropy shifts, while maintaining high throughput (44.89 TPS). These results validate that block-wise decoding ensures an improved decoding order, parallel decoding enables effective acceleration, and adaptive partition unlocks additional accuracy gains through constituent-aware boundary detection. Applying cumulative strategies, Swordsman achieves superiority in both inference speed and generation quality.

\begin{table}[t]
  \centering
  \caption{Efficiency comparison of decoding strategies along the DLM decoding evolution.}
  \label{decodingablation}
  \footnotesize                 
  \setlength{\tabcolsep}{3pt}   
  \renewcommand{\arraystretch}{1.05} 
  \resizebox{\linewidth}{!}{
  \begin{tabular}{@{}lccc cc@{}}
    \toprule
    \textbf{Method} & \textbf{Block-wise} & \textbf{Parallel} & \textbf{Adaptive} & \textbf{Accuracy} $\Uparrow$ & \textbf{TPS} $\Uparrow$ \\
    \midrule
    LLaDA     & $\times$ & $\times$ & $\times$ & 29.04 & 3.32 \\
              & $\checkmark$ & $\times$ & $\times$ & 77.40 & 8.39 \\
              & $\checkmark$ & $\checkmark$ & $\times$ & 77.56 & 42.36 \\
    \midrule
    Swordsman & $\checkmark$ & $\checkmark$ & $\checkmark$ & \textbf{81.43} & \textbf{44.89} \\
    \bottomrule
  \end{tabular}
  }
\end{table}

\subsection{Ablation Studies}
We conduct extensive experiments to understand how the overall strategies of Swordsman contribute to performance. Unless otherwise stated, all ablation studies are conducted on GSM8K using LLaDA-8B-Instruct, with caching disabled and a maximum generation length of 512 tokens.

\begin{figure}[H]
\centering
\includegraphics[width=\linewidth]{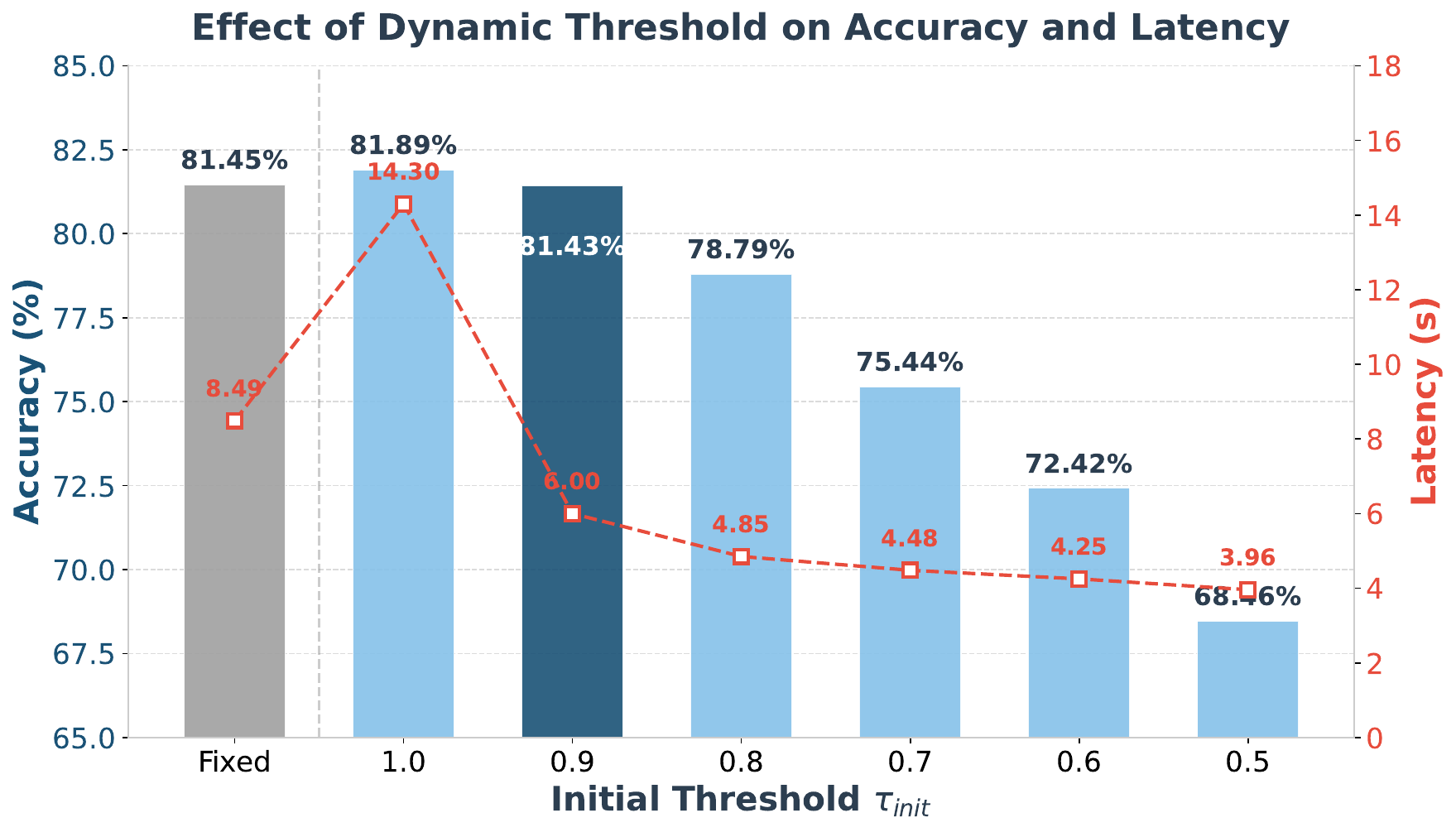}
\caption{Accuracy and latency comparison between fixed and dynamic thresholds with varying $\tau_{\text{init}}$} 
\label{adaptive_threshold}
\end{figure}

\paragraph{Effect of Adaptive Threshold}
We ablate the dynamic threshold mechanism in Figure \ref{adaptive_threshold}.  The fixed threshold $\tau_{fixed} =0.9$ (gray bar) enforces a rigid confidence constraint throughout the decoding process. When remaining tokens fail to meet $\tau_{fixed}$, the model falls into a degradation state, where tokens must be decoded sequentially by selecting the highest-confidence token one at a time, significantly slowing down inference speed. Our dynamic threshold adaptively relaxes confidence as tokens are decoded and information accumulated, enabling more parallel decoding. The optimal setting $\tau_{init} = 0.9$ achieves comparable accuracy (81.43\% vs. 81.45\%) while reducing latency by 2.49s. Lower ones (e.g., 0.5) further reduce latency but cause significant accuracy drops due to premature decoding of uncertain tokens. Conversely, $\tau_{init} = 1.0$ is overly strict, resulting in $\sim$
2.4× higher latency compared to the optimal trade-off point where $\tau_{init} = 0.9$ from excessive sequential decoding.

\section{Conclusion}
We introduce Swordsman, a training-free framework that achieves constituent-aware block-wise decoding through entropy-driven adaptive partition. By aligning blocks with semantic boundaries via entropy shift detection and modulating decoding strategies through dynamic thresholding, Swordsman addresses the semantic misalignment inherent in fixed block approaches. Experiments show Swordsman improves vanilla LLaDA accuracy from 77.40\% to 81.50\% on GSM8K with 8.79× speedup, and achieves +8.31\% over Fast-dLLM on HumanEval (35.59\% → 43.90\%) at matched speed. These results validate that entropy-based semantic awareness enables efficient parallel generation. Current limitations include validation primarily on block-wise DLMs and the potential need for dataset-specific hyperparameter tuning. Future work includes extending to semi-autoregressive models, developing adaptive parameter selection, and improving efficiency through learned mechanisms.

\section*{Impact Statement}
This paper presents work whose goal is to advance the field of Machine
Learning. There are many potential societal consequences of our work, none of which we feel must be specifically highlighted here.

\clearpage
\newpage

\bibliography{example_paper}
\bibliographystyle{icml2026}


\end{document}